\relax
\documentclass[letterpaper]{article} 
\usepackage{aaai22}  
\usepackage{times}  
\usepackage{helvet}  
\usepackage{courier}  
\usepackage[hyphens]{url}  
\usepackage{graphicx} 
\usepackage{caption} 
\DeclareCaptionStyle{ruled}{labelfont=normalfont,labelsep=colon,strut=off} 
\usepackage{algorithm}
\usepackage{algorithmic}
\usepackage{lineno}
\usepackage{amsmath}
\usepackage{amsfonts}

\usepackage{newfloat}
\usepackage{listings}
\floatstyle{ruled}
\newfloat{listing}{tb}{lst}{}
\floatname{listing}{Listing}
\title{Don't do it: Safer Reinforcement Learning With Rule-based Guidance}
\author{
    Ekaterina Nikonova,\\
    Cheng Xue,
    Jochen Renz
}
\affiliations{
    School of Computing\\ 
    The Australian National University\\ 
    Canberra, Australia\\
    ekaterina.nikonova, cheng.xue, jochen.renz@anu.edu.au
}

\usepackage{bibentry}

\begin{document}

\maketitle

\begin{abstract}
During training, reinforcement learning systems interact with the world without considering the safety of their actions. When deployed into the real world, such systems can be dangerous and cause harm to their surroundings. Often, dangerous situations can be mitigated by defining a set of rules that the system should not violate under any conditions. For example, in robot navigation, one safety rule would be to avoid colliding with surrounding objects and people. In this work, we define safety rules in terms of the relationships between the agent and objects and use them to prevent reinforcement learning systems from performing potentially harmful actions. We propose a new safe epsilon-greedy algorithm that uses safety rules to override agents' actions if they are considered to be unsafe. In our experiments, we show that a safe epsilon-greedy policy significantly increases the safety of the agent during training, improves the learning efficiency resulting in much faster convergence, and achieves better performance than the base model.
\end{abstract}

\section{Introduction}
In reinforcement learning, the agent tries to maximize the long-term return and often explores the world in a random manner without considering the safety of its actions. While safety is usually less important in simulated environments, in the real world, safe interaction is often a rather expected and necessary quality of the agent. For example, for a delivery robot it is important to not collide with any surrounding obstacles damaging itself or the package; for a self-driving car making safe decisions is crucial for the safety of the driver and others. In AI safety, this particular issue is known as \textit{safe exploration} \cite{Amodei2016ConcretePI} and has been widely explored in reinforcement learning \cite{Garca2015ACS, Hans2008SafeEF, Thomas2021SafeRL, Kobelrausch2021CollisionFreeDR, Pecka2014SafeET, Leike2017AISG, Gu2022ARO}. 

Multiple approaches have been taken by the researchers to provide safer reinforcement learning. For example, transforming optimization criteria \cite{Geibel2005RiskSensitiveRL, Howard1972, Ltjens2019SafeRL} or providing external guidance \cite{Garca2012SafeEO, Chow2018ALA, Clouse1992ATM, Geramifard2013IntelligentCC}.

Often, safe exploration and interaction can be ensured by stating safety rules for the particular domain. For example, consider the earlier mentioned real-world task for the delivery robot to navigate the streets without crashing into surrounding objects and people. In this task, the most important safety rule to follow is to avoid performing the action that would result in an immediate crash. This rule can be easily defined using spatial relationships between the robot and other objects, for example using qualitative spatial representation \cite{Cohn2008QualitativeSR, Clementini1997QualitativeRO}. In words, we can define a simple rule for this task as "if the object is in close proximity to the agent and the intended action would result in a collision - do not perform such action".

\begin{figure*}[h]
\centering
\includegraphics[scale=0.24]{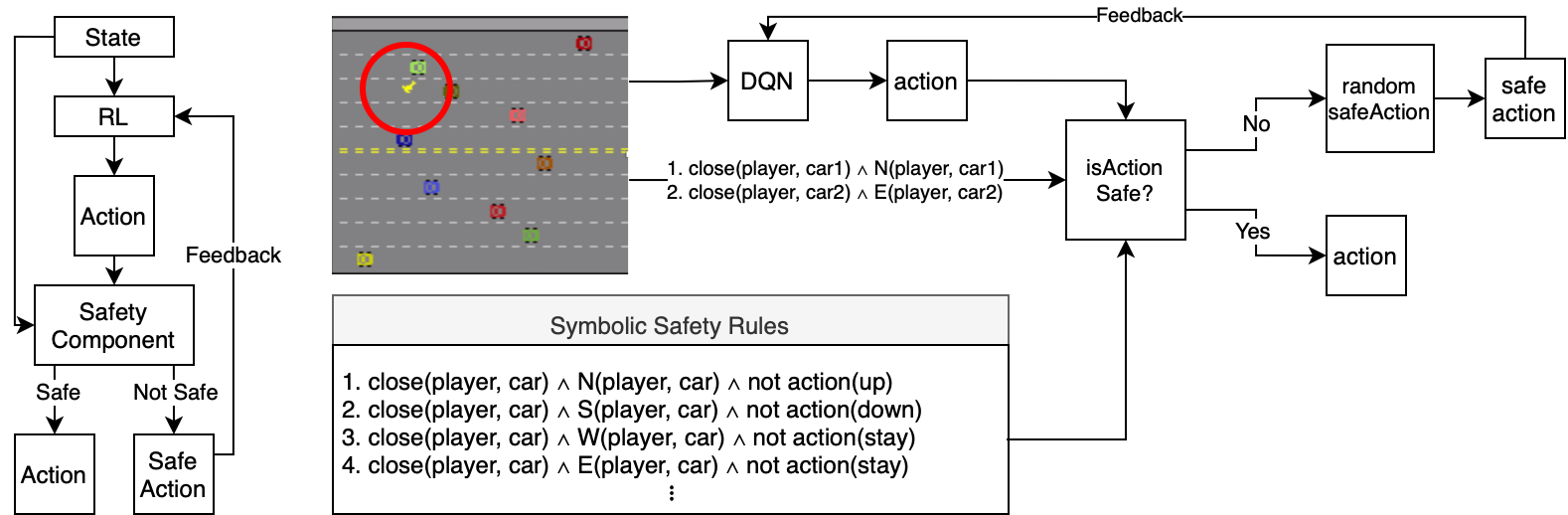}
\caption{(Left) General framework to validate the safety of the agent's decision using safety component. (Right) An example of framework implementation in Freeway. Here predicting action "up" would result in violating rule 1 and such a decision would be overwritten by safer action "down".}
\label{fig:framework}
\end{figure*}

The idea of stating rules to prevent agents from doing something unsafe goes a long way back to Issac Asimov's ideas in his "Three Laws of Robotics" \cite{Asimov}. In some cases, such basic safety rules are either known and can be injected as prior knowledge to the agent, or can be learned by the agent as it explores the world. 

In reinforcement learning, researchers have explored the idea of using prior knowledge to address \textit{safe exploration problem} \cite{Garca2015ACS, Pecka2014SafeET, Leike2017AISG}. One possible way to use prior knowledge is by using \textit{imitation learning} \cite{Santara2018RAILRI}. For example, the agent could imitate human demonstrations to avoid performing unsafe actions \cite{Ghavamzadeh2016SafePI}. Another approach is to include the safety constraints directly in the policy optimization \cite{Achiam2017ConstrainedPO}. Saunders proposed to entirely override the agent's actions if it was seen to be unsafe \cite{Saunders2018TrialWE}. We find Saunders's idea to be the most similar to ours, however, they use human oversight to override agent decisions and learn intervention policy from human demonstrations. While shown to be effective and safe, their approach requires a human to be in the loop which is both time-consuming and often unrealistic. In our work, we extend this idea and eliminate the need of constant human oversight. Instead, we use either human-defined or machine-learned safety rules to override agents' actions if they are unsafe. To the best of our knowledge, we have not found a method that uses symbolic safety rules directly to prevent an agent from performing unsafe actions.

In this work, we focus on the domains that require agents to act safely to do a task well. We propose a general framework to integrate the safety component together with reinforcement learning and provide one possible implementation of that framework using symbolic rules (Figure \ref{fig:framework}). In our experiments, we use human knowledge about the internal dynamics of the environment to manually construct symbolic safety rules. To show that our framework can work with different safety components, we provide a short experiment to learn the rules as the agent explores the world. We propose a new method to incorporate safety rules into the reinforcement learning algorithm. Specifically, we define a safe-$\epsilon$-greedy policy that uses symbolic rules to validate the safety of actions and overrides them with safer actions if necessary. In all our experiments we use the same underlying deep learning model with identical hyperparameters to ensure a fair evaluation of our method. We demonstrate that restricting the agent in its actions improves the overall performance, increases the safety of the agent, and significantly improves the efficiency of the agent achieving super-human or state-of-the-art performance faster. In our experiments, we test our method in five domains: a toy-domain Crossroad, and four OpenAI Gym Atari 2600 environments: Freeway, Frostbite, Ms-Pacman, and Space Invaders \cite{Brockman2016OpenAIG}. 


\section{Background}
In this work we consider a deterministic Markov Decision Process (MDP) $M = (S,A,T,r,\gamma)$, where $S$ is the state space, $A$ is the action space, $T:S \times A \rightarrow S$ the transition function, $r:S \times A \rightarrow r$ is a reward function, and $y \in [0, 1)$ the discount factor. 

A \textit{policy} $\pi : S \rightarrow A$ determines which action to take in each state. Typically, the goal of reinforcement learning is to find a policy that maximizes the expected discounted reward and is therefore considered to be \textit{optimal}. 

The \textit{Q-function} $Q^{\pi}(s,a) = \mathbb{E}^{\pi}[\sum^{\infty}_{t=0} \gamma^{t}r_{t}|s_{0}=s, a_{0} = a] $ measures the performance of the agent assuming it starts in a state $s$, takes action $a$ and follows the policy $\pi$ afterwards. 

The \textit{Value-function} $V^{\pi}(s) = \mathbb{E}^{a \sim \pi(s)}[Q^{\pi}(s,a)] $ measures the overall value of the state. Same as with policy, those functions can be optimal: $Q^{*}(s,a) = max_{\pi} Q^{\pi}(s,a) $ and $V^{*}(s) = max_{\pi} V^{\pi}(s)$. 
Finally, the \textit{optimal policy} can be retrieved from $Q^{*}$ as follows:  $\pi^{*}(s) = argmax_{a} Q^{*}(s,a) $.

In deep reinforcement learning, $Q$-function can be approximated using a nonlinear function approximator such as a neural network $Q(s,a,\theta_i)$, where $\theta_i$ are the weights of the Q-network at the i-th iteration. However, when using a nonlinear function approximator together with reinforcement learning, it can become unstable or even diverge due to the following problems: a) the correlation in the sequence of observations, b) correlations between Q values and target values $ r_t + \gamma max_{a}Q(s_t,a) $ and c) having a policy that is extremely sensitive to changes of Q value. 

A deep Q-network (DQN) addresses the first problem by using \textit{experience replay}. Experience replay is implemented by storing and later randomly sampling the observations experienced by the agent. This technique removes the correlation between the sequences of the observations by randomizing the collected data. We define the experience as $e_t=(s_t,a_t,r_{t+1},s_{t+1})$, and experience set as $M=\{e_1,…,e_t\}$.

In order to address the second problem, the notion of target network was introduced which is then used to calculate the loss: $L_i(\theta_i) = \mathbb{E}_{(s,a,r,s^{'})\sim U(M)} [ (r+\gamma max_{a^{'}}Q(s^{'},a^{'},\theta_i^{-}) -Q(s,a,\theta_i ))^2 ]$, where $i$ is the iteration, $\gamma$ is a discount factor, $\theta_i$ are weights of so-called online Q-network and $\theta_i^{-}$ are weights of a target network or so-called offline Q-network. The target network is called offline since its weights are only updated every $C$ steps with a copy of online network weights, while the online network is updating every iteration $i$.


\section{Guidance Using Symbolic Rules}
\subsection{Definitions} 
\label{definitions}
\begin{figure}[h]
\centering
\includegraphics[scale=0.2]{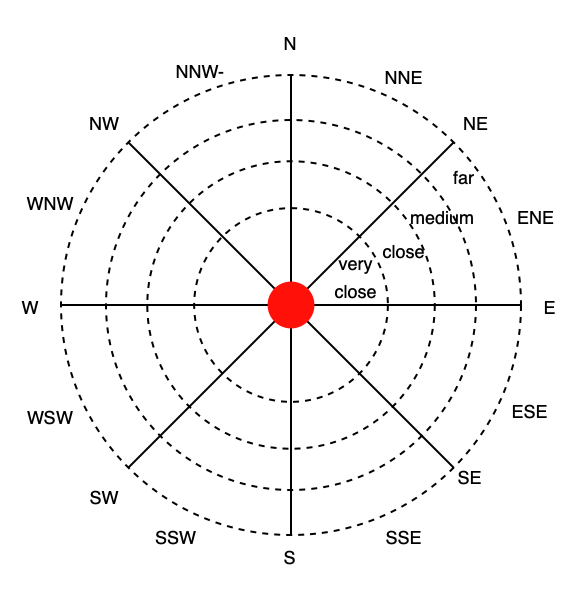}
\caption{Qualitative directional and distance representation used for the safety rules. }
\label{fig:qsr}
\end{figure}

\paragraph{Qualitative Spatial Representation}
In order to construct a symbolic representation of the states, we extract qualitative spatial relationships (QSR) between the objects and the agent. In particular we use cone-shaped directional representation \cite{Renz2004QualitativeDC} and a qualitative distance representation \cite{Frank1992QualitativeSR}. Figure \ref{fig:qsr} shows an example of a combination of directional and distance representations. We use this combination to construct safety rules and relational representation of the state.

\paragraph{Safety Rules} One obvious approach would be to define safety in terms of the state transitions and next states \cite{Hans2008SafeEF}. However, such a definition would require an agent to have a model which is not always possible. Instead, we look at the relationships between the objects to determine if the action is safe or not in the given state. We define a \textit{safety rule} to be a conjunction of n-ary relationships between objects and action which if satisfied would compromise safety: $safetyRule=(r_{1}(o_{1}, ... o_{n}) \wedge ... \wedge r_{m}(o_{1},...,o_{n}) \wedge \lnot action(a) )$, where $r_{i}$ is a QSR relationship, $o_{j}$ is an object and $action(a)$ is the action. For example, a safety rule would prevent agent from performing an action that could lead to the collision with the near-by objects (i.e. $close(agent, o) \wedge N(agent, o) \wedge \lnot action(up)$). We define \textit{safety rules} to be a collection of such rules. Figure \ref{fig:framework} (Right) demonstrates symbolic safety rules for Freeway.
\paragraph{Is the action safe?} We call an action $a_{t}$ to be \textit{safe} in state $s_{t}$, if performing that action would not violate any safety rule. To validate that action is safe, we extract symbolic relationships from the state $s_{t}$ as $s^{symb}_{t} = \{ r_{1}(o_{1}, ... o_{n}),...,r_{m}(o_{1},...,o_{n}) \}$, where $r_{i}$ is a QSR relationship and $o_{j}$ is an object in the state $s_{t}$. For example, we can extract spatial relationships between the nearby objects and the agent (i.e. $close(agent, car_{1}) \wedge N(agent, car_{1})$). We then conjugate it with the symbolic representation $a^{symb}_{t}$ of action $a_{t}$ (i.e. $action(a_{t})$). Finally, we define $isActionSafe(s_{t}, a_{t})$ =

\begin{equation}
 \label{eq:isActionSafe}
  \begin{cases}
 1,& \text{if } (s^{symb}_{t} \wedge \lnot a^{symb}_{t})\cap safetyRules = \emptyset \\
 0,& \text{otherwise}
 \end{cases}
\end{equation}

Where $s^{symb}_{t}$ are the relationships extracted from $s_{t}$ and $a^{symb}_{t}$ is symbolic representation of $a_{t}$. We note that in this work, we only extract spatial relationships between objects and agents that fall within a certain region around the agent (Figure \ref{fig:framework} (Right)). Consider Figure \ref{fig:framework} that demonstrates a framework implementation in a Freeway. Here the safety rules prevent the action that would result in a collision with a car. In this example the car is immediately above the agent (i.e. $s^{symb} = (close(agent, car_{1}) \wedge N(agent, car_{1}))$). If the DQN predicts action "up" (i.e $close(agent, car_{1}) \wedge N(agent, car_{1}) \wedge action(up)$), rule 1 would be violated, and such action would be overridden with a safe action. 

\begin{algorithm}
\caption{Guided Exploration}
\label{alg: guided-exploration}
\textbf{Input}: $Q$, $s_{t}$ \\
\textbf{Output}: $a$ \\
\begin{algorithmic}[1] 
\STATE n $\sim$ $\mathcal{U}_{[0, 1]}$
\IF{n \textless $\epsilon$}
   \STATE a $\gets$ \textit{selectRandomSafeAction}($s_{t}$)
\ELSE
    \STATE a $\gets$ $argmax_{a} Q(s_{t},a)$
\ENDIF
\RETURN a
\end{algorithmic}
\end{algorithm}

\paragraph{Random safe action} To select a random safe action in the state $s_t$, we define $selectRandomSafeAction(s_{t}) =$
\begin{equation}
 \label{eq:selectRandomSafeAction}
\begin{cases}
a \in A_{safe},& \text{if } A_{safe} \neq \emptyset \\
a \in A,& \text{otherwise}
\end{cases}
\end{equation}
In Equation \ref{eq:selectRandomSafeAction}, action is sampled from $A_{safe}$ or $A$ uniformly and $A_{safe}=\{a^{i}_{t} | isActionSafe(s_{t}, a^{i}_{t}) = 1, a^{i}_{t} \in A \}$ is a collection of all safe actions. The actions that would result in violating any safety rule are called to be \textit{unsafe} and should be avoided by the agent whenever it is possible.

\subsection{Safe $\epsilon$-greedy policy}
In this section, we propose a method to inject the safety rules as part of the modified, safer version of the $\epsilon$-greedy algorithm. \textit{Safe $\epsilon$-greedy} algorithm directly prevents the agent from performing unsafe action by overriding its decision with a safe action (if possible). We hypothesize that restricting the agent in its actions, will not only make learning safer but likely will reduce the size of the search space and therefore improve the performance. 

In each time-step, safe $\epsilon$-greedy leverages safety rules to determine the safety of agents' actions. We provide two variants of the safe $\epsilon$-greedy algorithm: Guided exploration and Full-guidance. First validates and overrides agents' actions during the exploration phase only, while second during the whole training process. We describe the algorithms further in the section.

Recall the $\epsilon$-greedy policy \cite{suttonrlintro}:
\begin{equation}
 \label{eq:epsilongreedy}
\pi(s_t) = 
\begin{cases}
a \in $A$,& \text{if } n \textless \epsilon, n \in \mathcal{U}_{[0, 1]} \\
argmax_{a} Q(s_{t},a),& \text{otherwise}
\end{cases}
\end{equation}
By definition, the epsilon-greedy policy (Equation \ref{eq:epsilongreedy}) selects a random action during the exploration phase, which, without any safety check, could result in the agent damaging itself or others. Instead, we propose to select a random safe action.

\begin{figure*}
\centering
\includegraphics[width=.1581\linewidth]{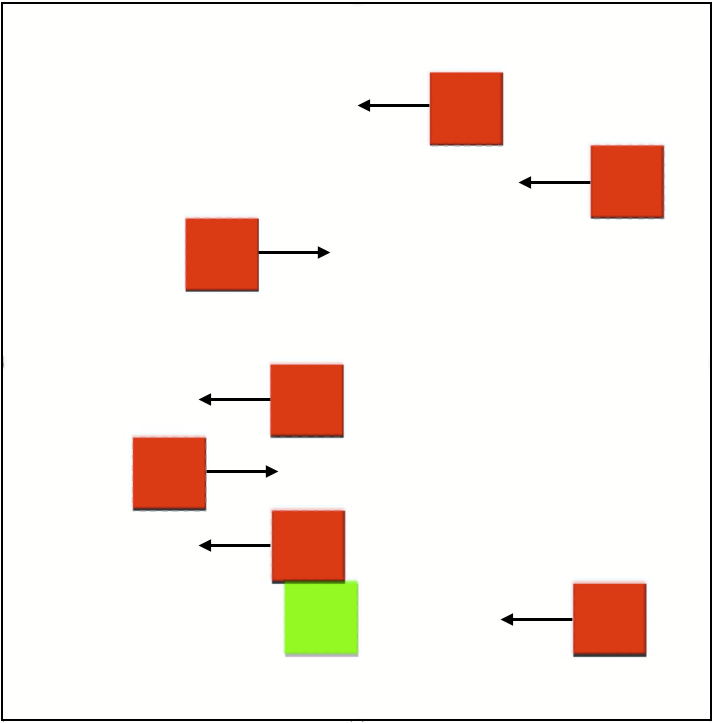}
\includegraphics[width=.207\linewidth]{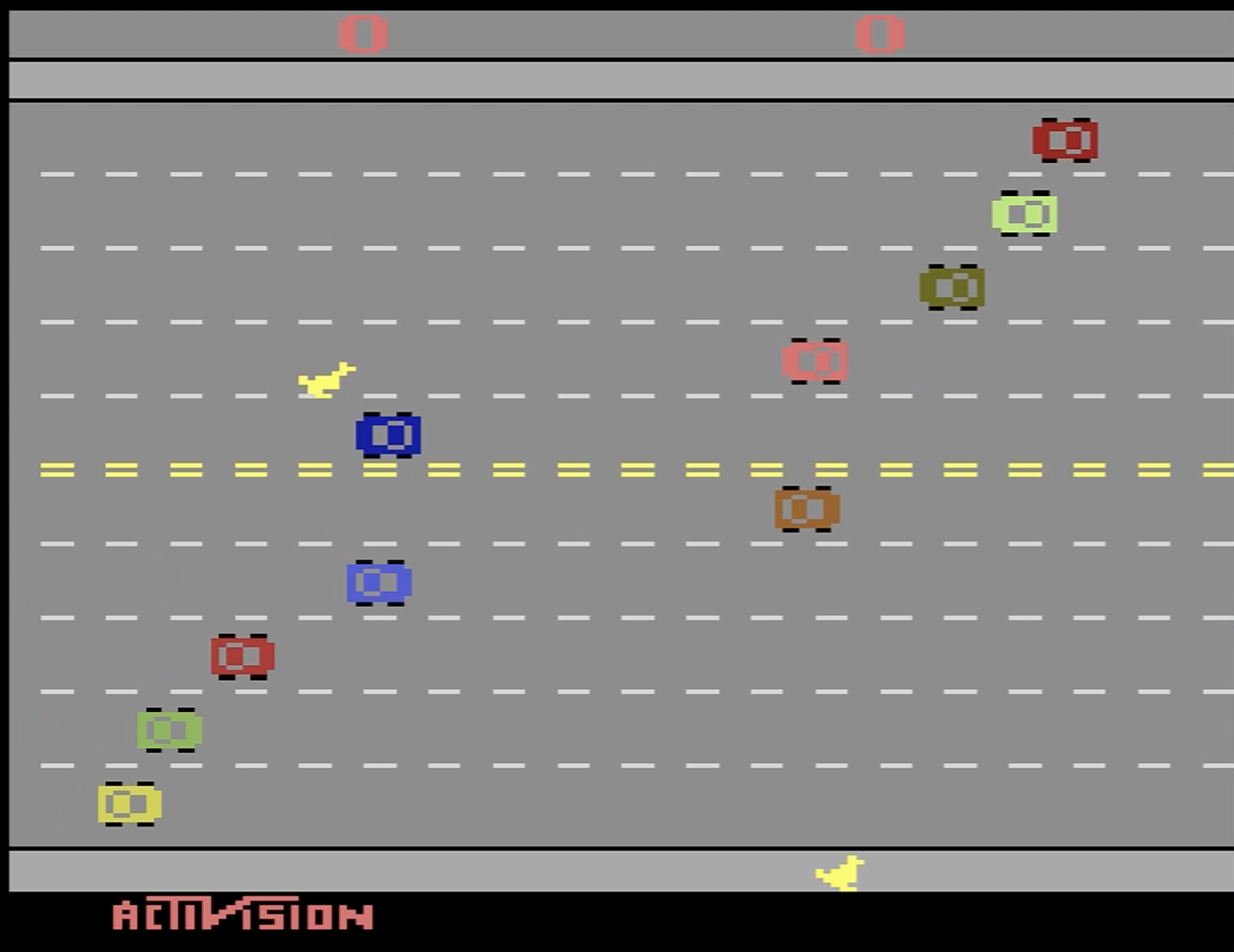}
\includegraphics[width=.18\linewidth]{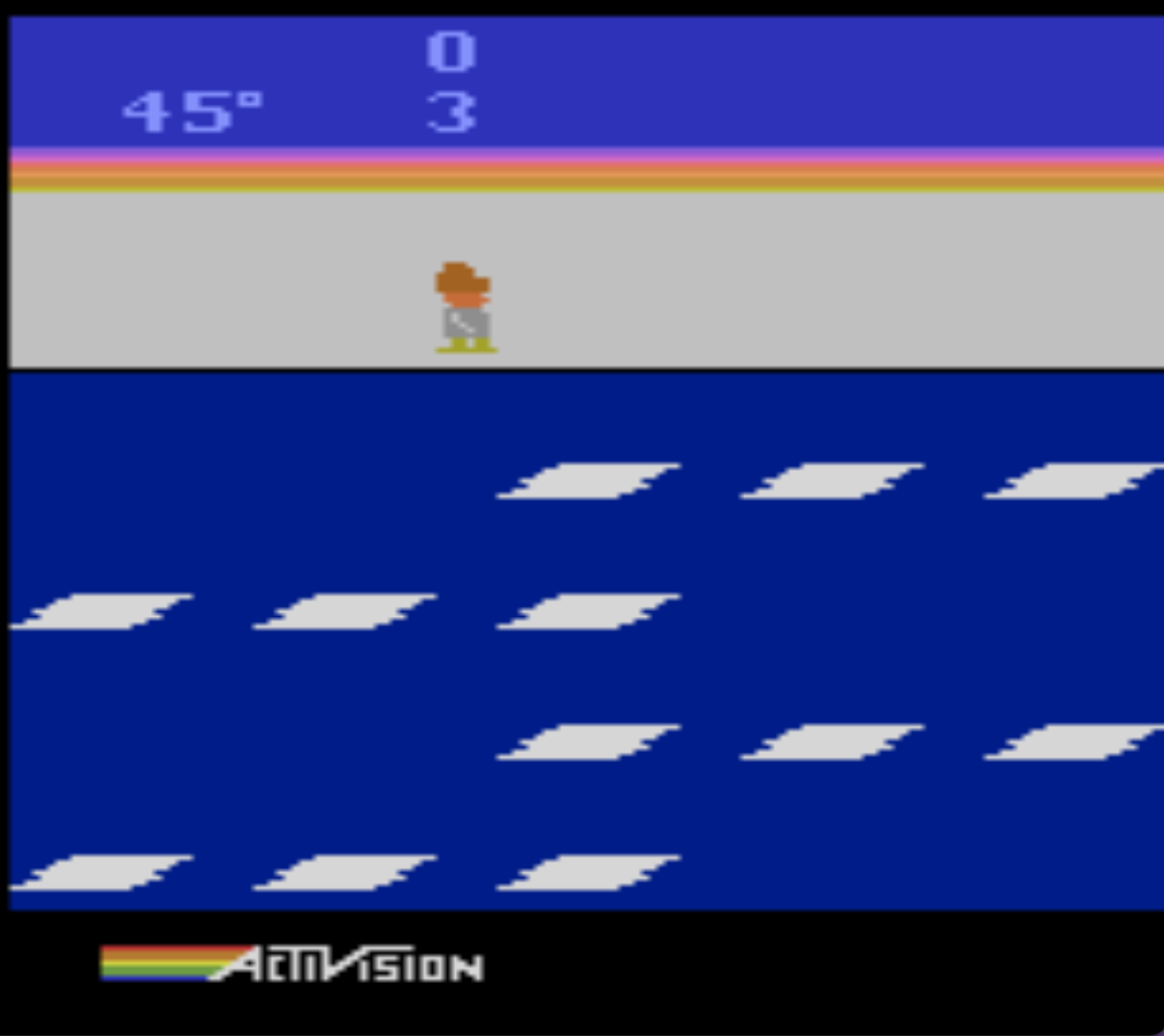}
\includegraphics[width=.196\linewidth]{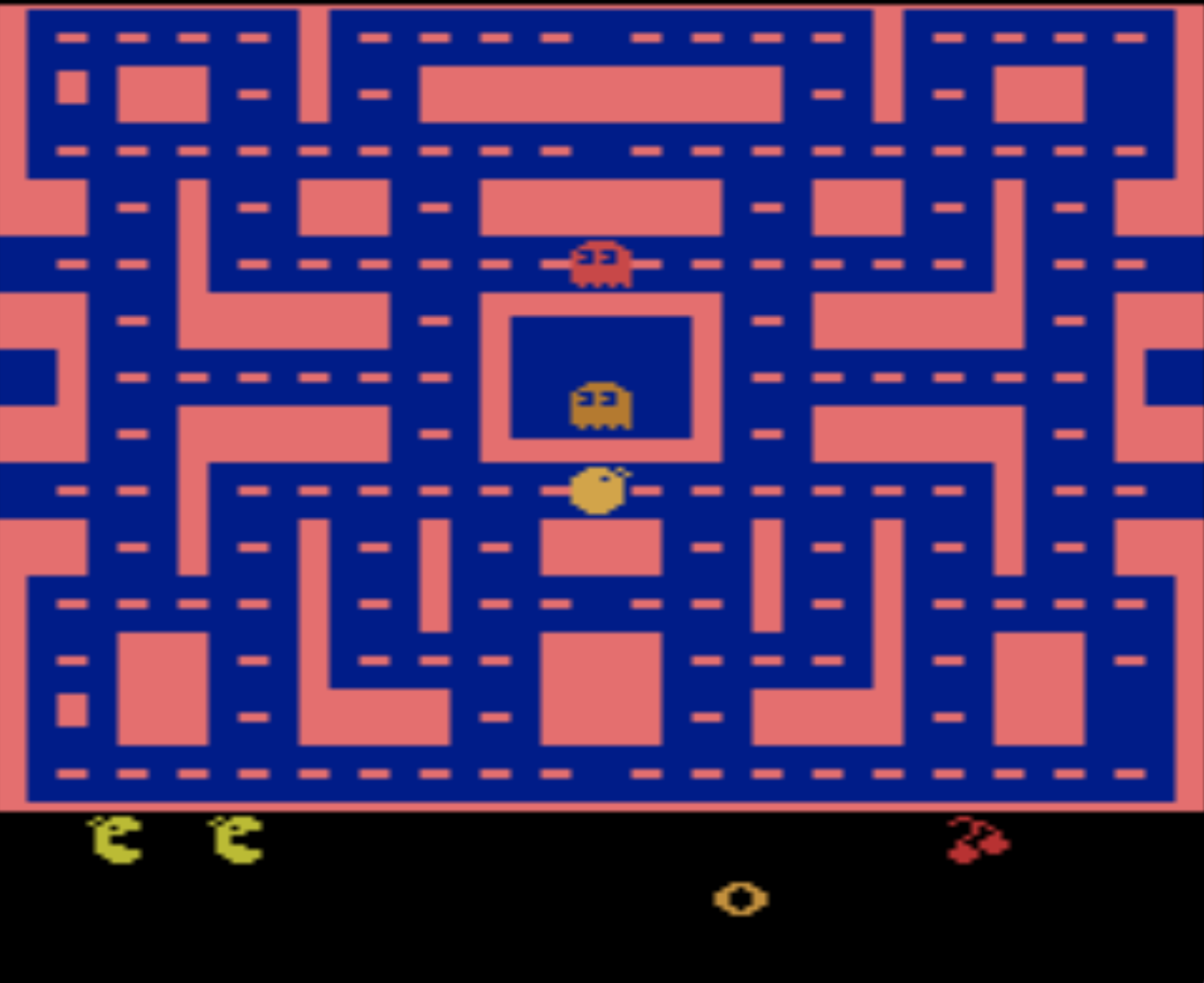}
\includegraphics[width=.186\linewidth]{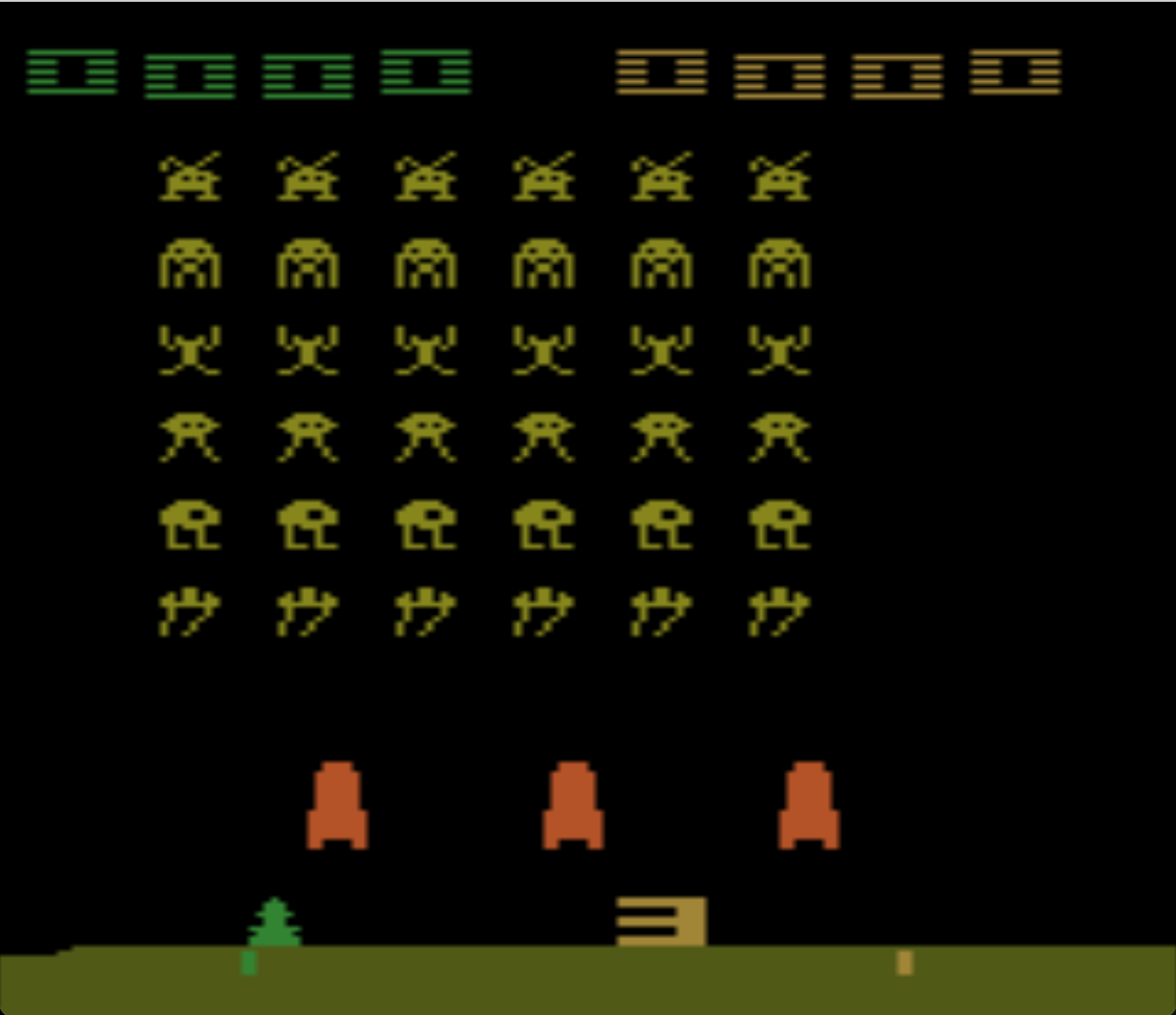}
\caption{Testing domains used in this work. From left to right: Crossroad, Freeway, Frostbite, Ms-Pacman, and Space Invaders.}
\label{fig:test-domains}
\end{figure*}



\begin{algorithm}
\caption{Full Guidance}
\label{alg: full-guidance}
\textbf{Input}: $Q$, $s_{t}$ \\
\textbf{Output}: $a$ \\
\begin{algorithmic}[1] 
\STATE n $\sim$ $\mathcal{U}_{[0, 1]}$
\IF{n \textless $\epsilon$}
    \STATE a $\gets$ \textit{selectRandomSafeAction}($s_{t}$)
\ELSE
    \STATE a $\gets$ $argmax_{a} Q(s_{t},a)$
    \IF{not \textit{isActionSafe}($s_{t}$, $a$)}
        \STATE a $\gets$ \textit{selectRandomSafeAction}($s_{t}$)
    \ENDIF
\ENDIF

\RETURN a
\end{algorithmic}
\end{algorithm}

Algorithm \ref{alg: guided-exploration} demonstrates how safety rules can then be embedded into the $\epsilon$-greedy policy resulting in its safer version. Here instead of selecting a random action, the algorithm selects random safe action. We note that safe action will be added to the experience replay $M$ (or Q-table as in tabular RL) to aid further training (feedback in Figure \ref{fig:framework}).

While Algorithm \ref{alg: guided-exploration} will ensure the agent is behaving safely during the exploration, it does not guarantee that once the exploration phase is over the agent will continue to behave safely. For example, if the agent encounters a previously unseen state, it could perform an unsafe action. Therefore, to ensure that the agent behaves safely during the entire training process, we must continue to supervise its actions.

Algorithm \ref{alg: full-guidance} presents an approach that provides the agent with guidance during the whole training process. Same as Algorithm \ref{alg: guided-exploration}, it will select safe actions during the exploration phase. However, in addition to it, when an action is selected using a learned $Q$-function, the algorithm will check whether this action is safe, and if not, it will randomly select a safe action. We hypothesize that such strict supervision of the actions performed by the agent will increase the safety of the agent. Ultimately, as the result of this algorithm, many unsafe actions will not be explored thus reducing the size of the search space. One could note that theoretically doing something unsafe could be a necessary step to achieve maximum performance, however here we prioritize safety over reward.

\section{Experiments}
\paragraph{Testing Domains}
In this work, we use five testing domains: Crossroad, and four OpenAI Gym Atari 2600 domains: Freeway, Frostbite, Ms-Pacman, and Space Invaders (Figure \ref{fig:test-domains}). 

Crossroad is an initial testing domain inspired by the Atari 2600 Freeway game. It has 7 cars moving horizontally either left or right and discrete action space with the ability to move left, right, down, and up or stay in place. The goal of the game is to cross the road without colliding with a car. Colliding with a car will result in the immediate end of the game. The agent is given the reward of 1 if the game is won, -1 if it has collided with a car, and 0 otherwise. 

Atari 2600 Freeway, is a game similar to Crossroad with cars moving left or right horizontally. However, in comparison to Crossroad, agents can only move up and down or stay in place. The goal of the game is to cross the road the maximum possible number of times before the timer runs out. Here colliding with a car will not result in the end of the game. The agent is given the reward of 1 if it crosses all roads and 0 in any other case. In Atari 2600 Frostbite, the agent needs to jump onto the platforms to build an igloo while avoiding falling into the water. In Atari 2600 Ms-Pacman agent needs to collect all dots while avoiding the ghosts (however, ghosts can be eaten in some cases for extra reward). Lastly, in Atari 2600 Space Invaders the agent needs to shoot enemies while dodging the bullets.




We chose these five domains as to win and master those games, the agent needs to act safely. For example, in Crossroad and Freeway agent needs to avoid the cars, in Ms-Pacman it should avoid ghosts, in Frostbite it should not jump into the water and in Space Invaders, it should avoid the bullets. All of this can be reformulated to real-world tasks such as avoiding obstacles and not falling off from high places. In addition to those five domains, our method can be easily applied to any other domain with well-defined safety rules. 

\paragraph{Agents and Settings} 
We evaluate safe $\epsilon$-greedy policy in tabular and deep reinforcement learning settings. As representation learning is not a focus of this work, we use symbolic state representation for the Crossroad and RAM for Atari 2600. In order to extract symbolic representation for the Atari games, we use the AARI package \cite{Anand2019UnsupervisedSR}. We note, however, that this part can be easily replaced with any state-of-the-art image processor to extract the location of the objects. We test two variants of the safe $\epsilon$-greedy policy using Guided-exploration and Full-guidance algorithms.

Tabular Q-learning allows us to measure the impact of the method without complications and instability of deep reinforcement learning. For tabular Q-learning, we test three agents: vanilla Q-learning, Q-learning + Guided Exploration, and Q-learning + Full-guidance. 

\begin{figure}[h]
\centering
\includegraphics[scale=0.1]{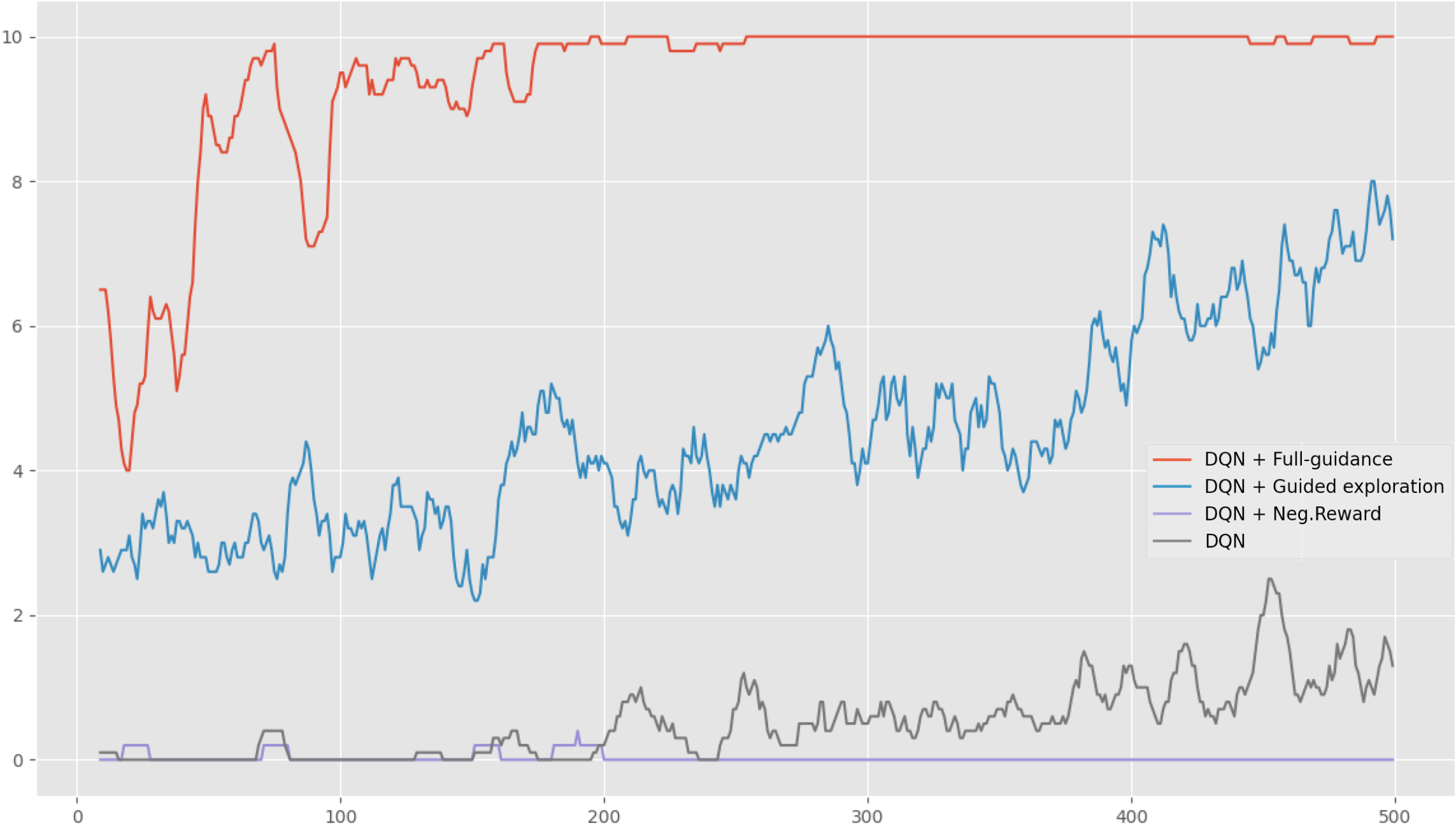}
\caption{Average rewards of DQN agents in Crossroad with 0-1 reward.}
\label{fig:crossroad-rewardshaping}
\end{figure}

For deep reinforcement learning experiments, we use Deep Q-network as a base model due to its relative simplicity and widespread in the research community. However, our method can be used with any other off-policy algorithm as a substitute for $\epsilon$-greedy policy. We test three agents: vanilla DQN, DQN + Guided Exploration (DQN+GE), and DQN + Full-guidance (DQN+FG). We note that we use the same hyperparameters and model architecture for all three of our tested agents. The only difference between the three agents is in action-selection algorithm they use during training (either Equation \ref{eq:epsilongreedy}, or Algorithms \ref{alg: guided-exploration} and \ref{alg: full-guidance}).

We compare the performance of the agents in Atari games to the performance of humans, random agent \cite{Lazaridis2020DeepRL}, and image-based DQN (image-DQN) that was trained for 50 million frames \cite{mnih2015humanlevel}. We include the results of image-based DQN to show the difference between the RAM and image state representation.

\paragraph{Safety Rules for Testing Domains}
As mentioned earlier, safety rules can be defined in terms of the relationships between the objects and actions. In this work, we define symbolic safety rules for each testing domain manually. However, we note that manually defining the rules is both cumbersome and time-consuming \cite{Weld1994TheFL}. To address this, we hypothesize that those rules can be learned autonomously. Later in the paper (in "Learning Rules Experiment" section), we present a short experiment showing that: a) rules can be learned and b) using learned rules will work with our method as well.

In both Crossroad and Freeway the agent should not collide with the cars. We define safety rules in terms of relationships between the cars and the agent (Figure \ref{fig:framework} (Right)). For example, to determine whether the action up is safe we check whether there is a car directly above the agent. If not then it is safe to move up. Similar rules apply to all other directions and actions.

In Frostbite the agent should avoid falling into the water. Similarly, safety rules can be defined in terms of the relationships between platforms and the agent. For example, to determine whether the action "down" is safe, we check whether there is a platform beneath the agent.

\begin{figure}[h]
\centering
\includegraphics[scale=0.28]{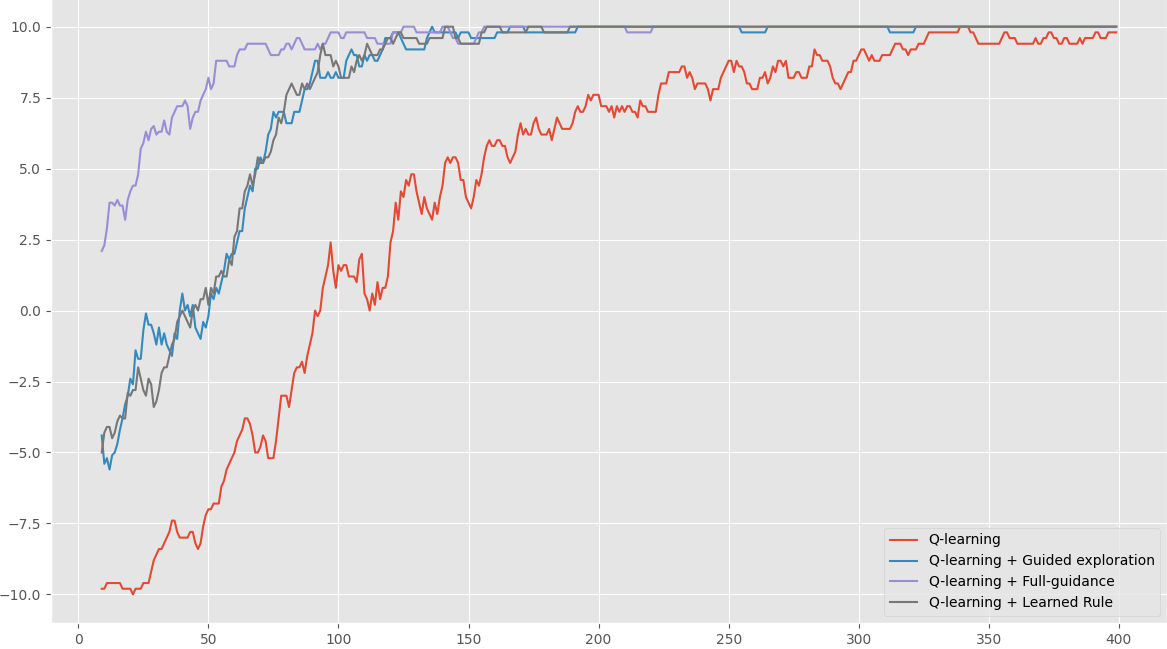}
\caption{Average rewards of Q-learning agents and learning rule agent in Crossroad.}
\label{fig:crossroad-learnedrule}
\end{figure}

In Ms-Pacman, the rule is quite similar to Crossroad and Freeway - to not collide with the ghosts, with the exception that sometimes ghosts can be eaten to receive bonus points. As previously this rule can be defined in terms of relationships between the game objects.

Lastly, in Space Invaders, the agent needs to avoid the bullets, which can be defined in terms of the relationships between the bullets and the player. When the bullet is close to the player, the player needs to move to dodge it. We provide full safety rules for all domains in the Appendix.





\paragraph{Difference to the Reward Shaping}
One logical question that one may have is whether constraining the agent in its action will lead to similar performance as providing the negative rewards for such actions. One obvious downside of this approach is that a negative reward will not completely prevent the agent from performing unsafe actions. The agent would still need to perform an action to learn that it is unsafe. Often, in the real world, this would be too much of a risk. We hypothesize, that since the agent would still need to explore the larger search space, likely it will not outperform the safe $\epsilon$-greedy agents. 

Figure \ref{fig:crossroad-rewardshaping} provides the learning curves of DQN agents in Crossroad. Originally, as for example in Figure \ref{fig:learning-curves}, Crossroad provides the reward of -1 if the agent collides with the car, 1 if it crosses all roads, and 0 in all other cases. However, we modified the reward function for this experiment to give a reward of 1 for crossing all roads and 0 otherwise. DQN+Neg.Reward is the only agent that was receiving a negative reward when colliding with a car, thus emulating the scenario where human shapes the reward function for each domain. In this experiment, the agent with a negative reward did not outperform the safe $\epsilon$-greedy agents nor vanilla DQN with 0-1 reward.

\paragraph{Learning Rules Experiment}
One obvious limitation of using human-defined safety rules is that it takes time to define and implement them \cite{Weld1994TheFL}. However, we hypothesise that such rules can be learned by agents automatically and conduct a short experiment to test this hypothesis. We demonstrate that learned rules can boost efficiency similarly to manually defined rules. We note, that learning rules is beyond the focus of this work and leave the full-scale experiment to future work.

\begin{figure}[h]
\centering
\includegraphics[scale=0.235]{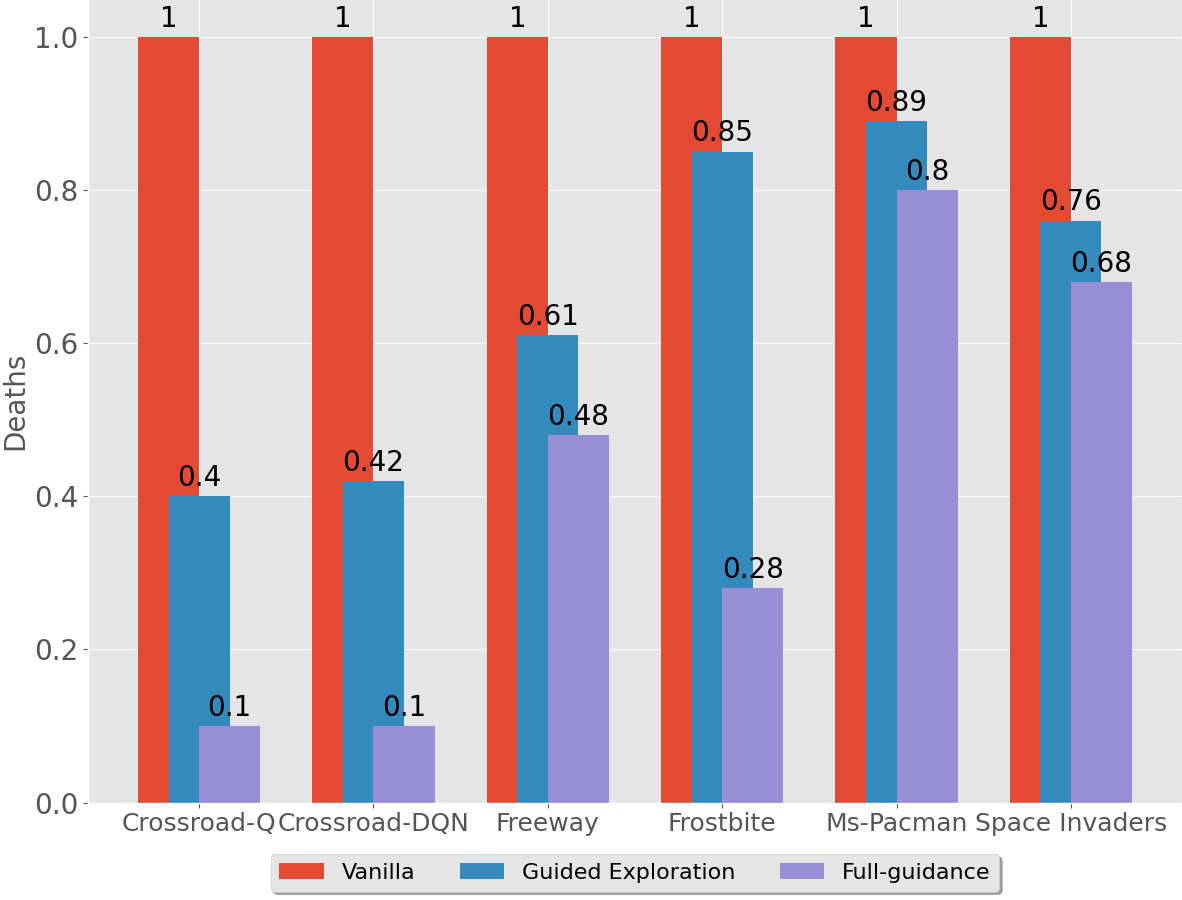}
\caption{The scaled number of times each agent has either died or collided relatively to the base model such as DQN or Q-learning. The lesser the number the safer training is (meaning fewer collisions or deaths).}
\label{fig:all_bumps}
\end{figure}

For this experiment we use supervised learning as safety component in our framework (Figure \ref{fig:framework} (Left)). We train logistic regression on the collection of symbolic states and classify them to "safe" and "unsafe" categories. In each state $s_{t}$ we use the model to get the next state $s_{t+1}$ following the action $a_{t}$ and use classifier to determine whether the $s_{t+1}$ is safe. We provide full details on the classifier and the approach in the Appendix. Figure \ref{fig:crossroad-learnedrule} shows the learning curves of the four agents: vanilla Q-learning, Q-learning with learned rule and Guided exploration, and two agents that use human-defined rules: Q-learning + Guided exploration, and Q-learning + Full-exploration. We conclude that using a learned rule is virtually almost identical to human-defined rules. This concludes that rules can be learned and used in safe $\epsilon$-greedy policy, we leave the full-scale experiment with other domains for future work.

\section{Results and Discussions}
In this subsection, we present the results of the experiments in five domains. We discuss how a safe $\epsilon$-greedy policy increases the safety of the agent during training, improves the learning efficiency, and achieves better performance than the base models.

\paragraph{Increased Safety}
Safe $\epsilon$-greedy leverages the safety rules to override agents' actions if they are considered to be unsafe. We hypothesize that this would decrease the number of collisions or deaths during training and potentially help the agent to learn a safer policy. Figure \ref{fig:all_bumps} shows the relative number of times each agent has died (or collided) during training. In all domains safe $\epsilon$-greedy agents demonstrated a noticeable reduction in the number of deaths. Based on those results, we hypothesize that in the real world our method would reduce the potential cost of training as the agent would damage itself fewer times. We note that in some domains the decrease in the deaths was not as significant as in other domains. We hypothesise that this is due to the short horizon of the safety rules. In some domains (i.e. Ms-Pacman), agent needs to plan its actions several steps ahead to avoid states where death is inevitable. One possible way to deal with it would be to use a more sophisticated safety component to validate the safety of the agents' actions. 

\begin{figure}[h]
\centering
\includegraphics[scale=0.25]{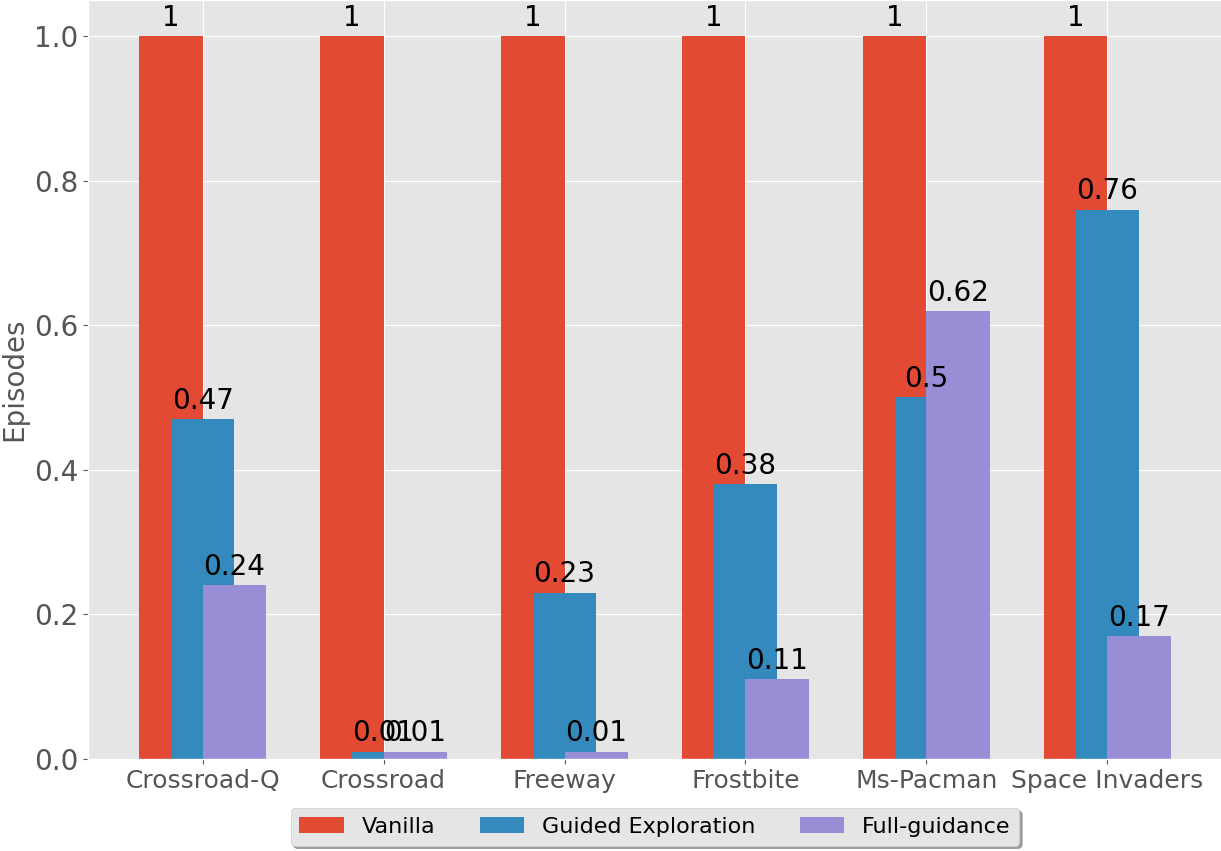}
\caption{The number of episodes it took each agent to achieve the maximum performance of the vanilla DQN or Q-learning. The lesser the number the more efficient the agent was (meaning faster convergence).}
\label{fig:rel_performance}
\end{figure}

\begin{figure*}[h]
\centering
\includegraphics[width=.33\linewidth]{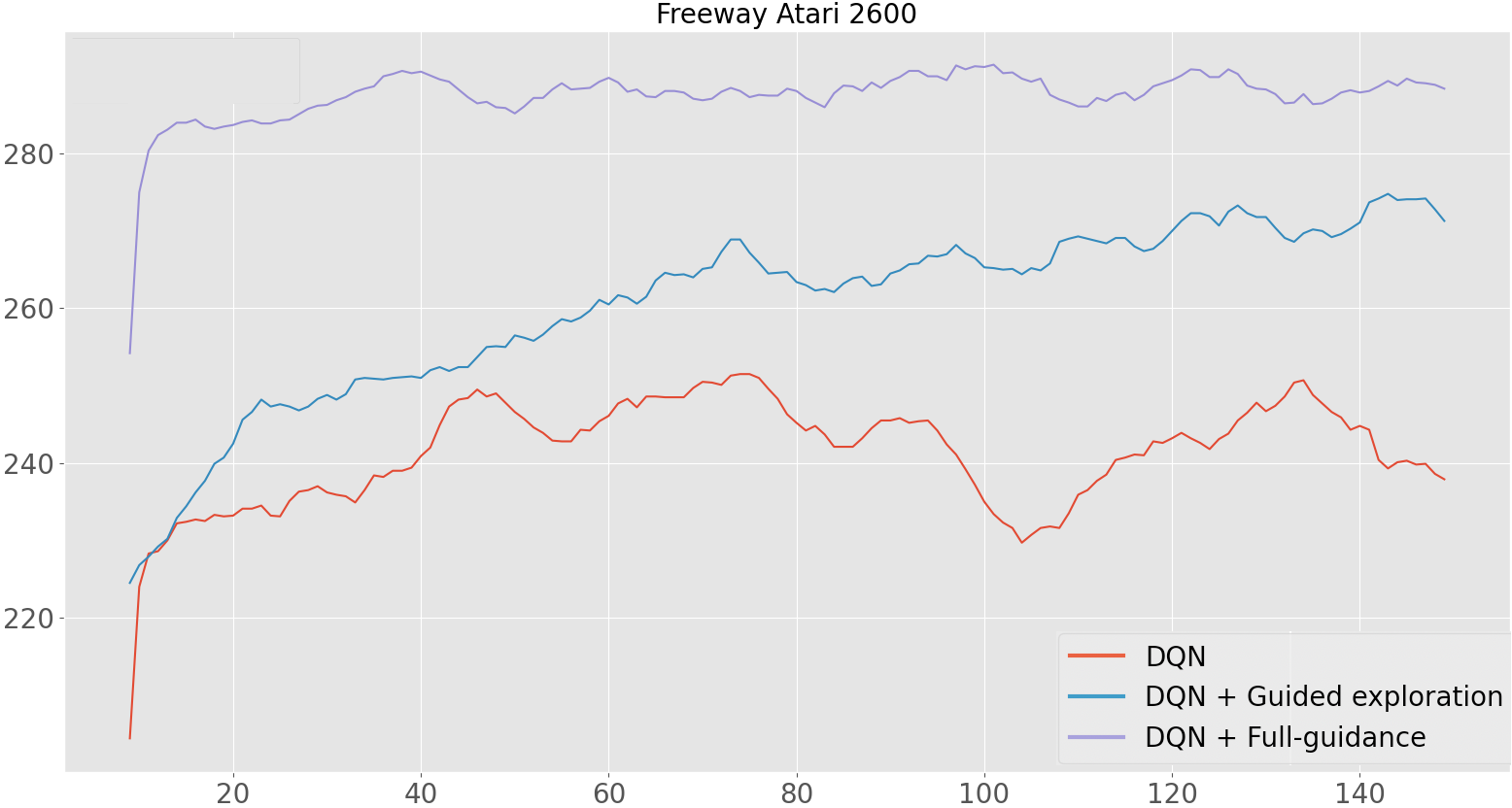}
\includegraphics[width=.33\linewidth]{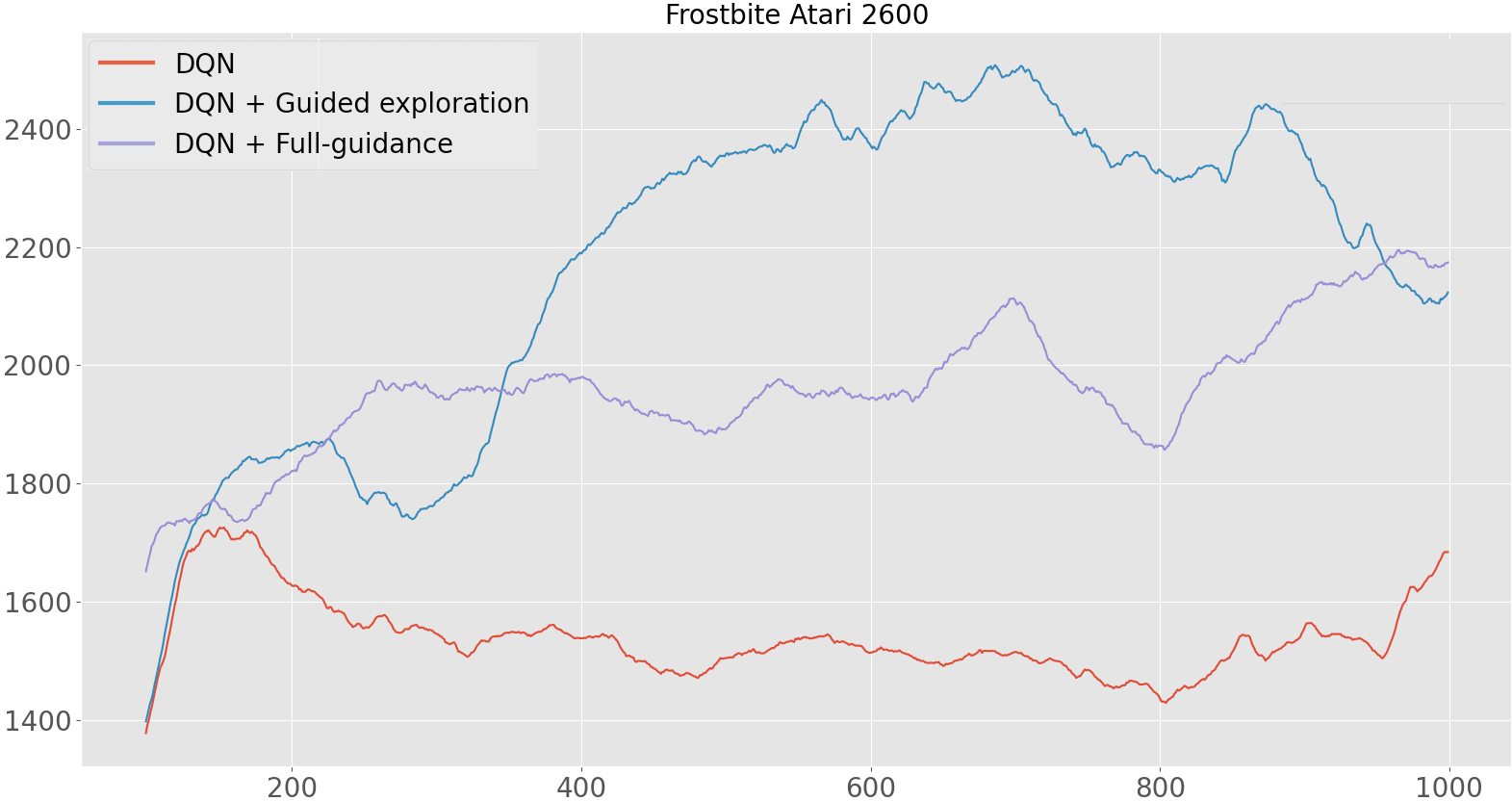}
\includegraphics[width=.33\linewidth]{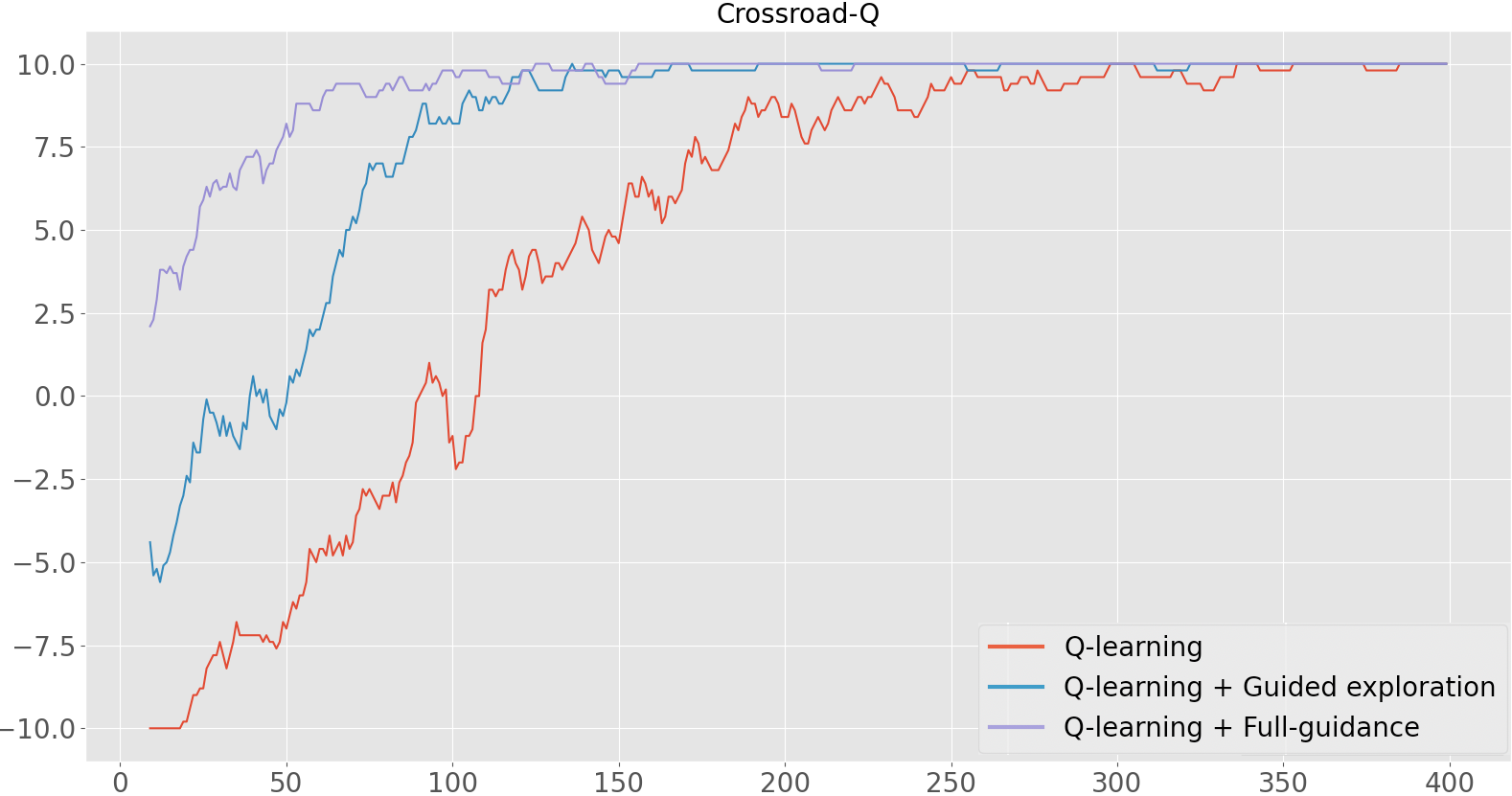}
\includegraphics[width=.33\linewidth]{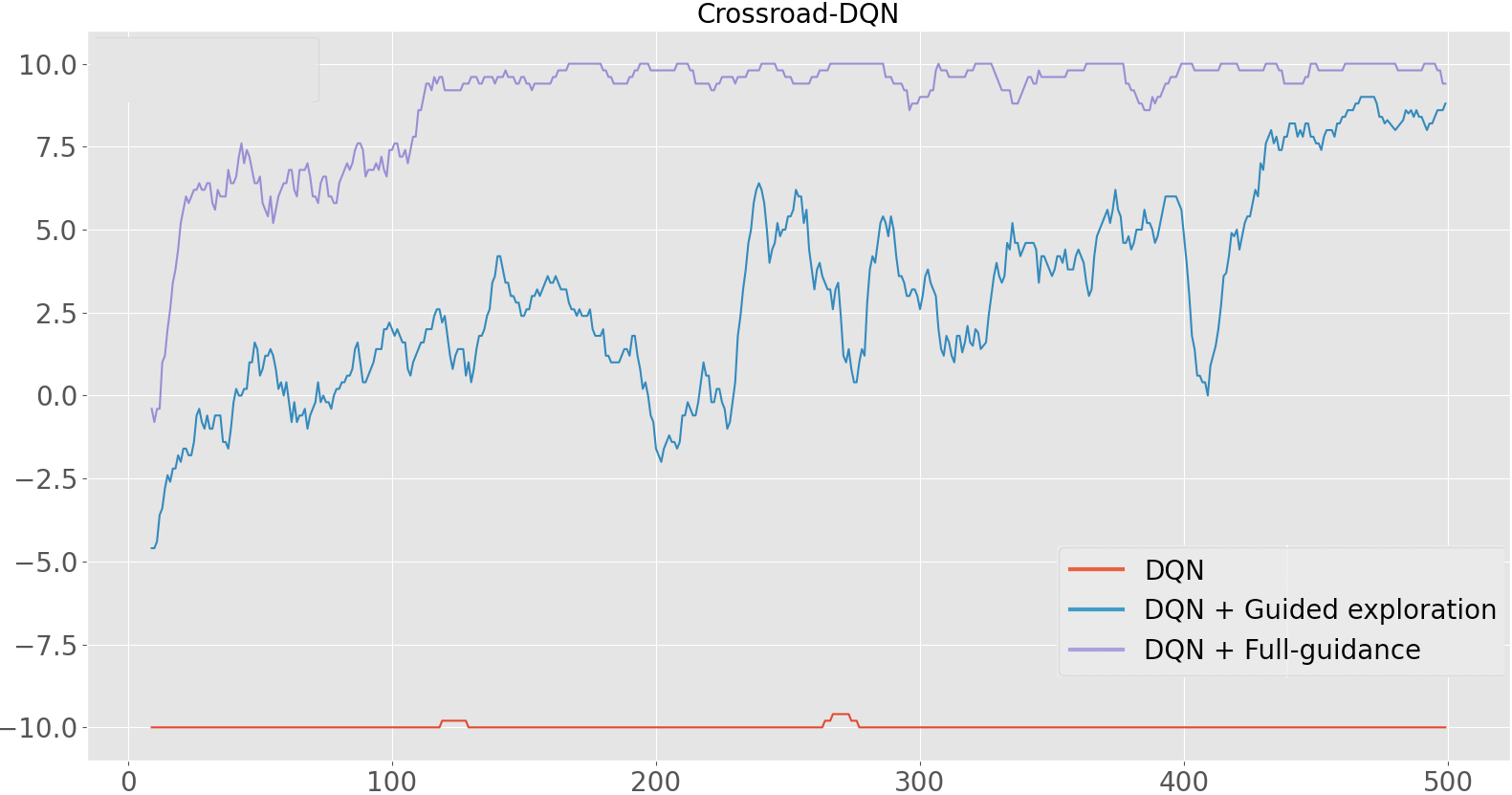}
\includegraphics[width=.33\linewidth]{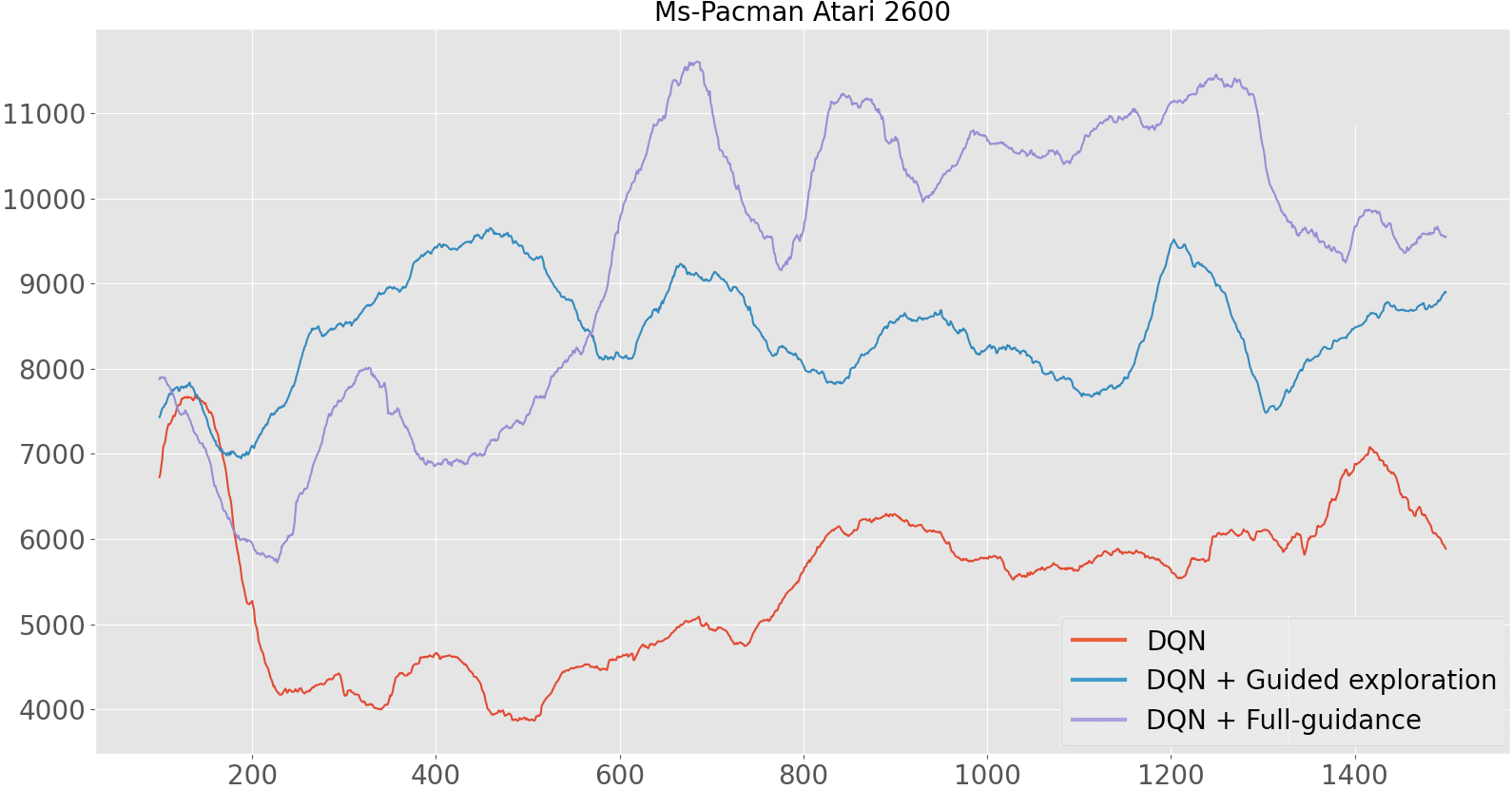}
\includegraphics[width=.33\linewidth]{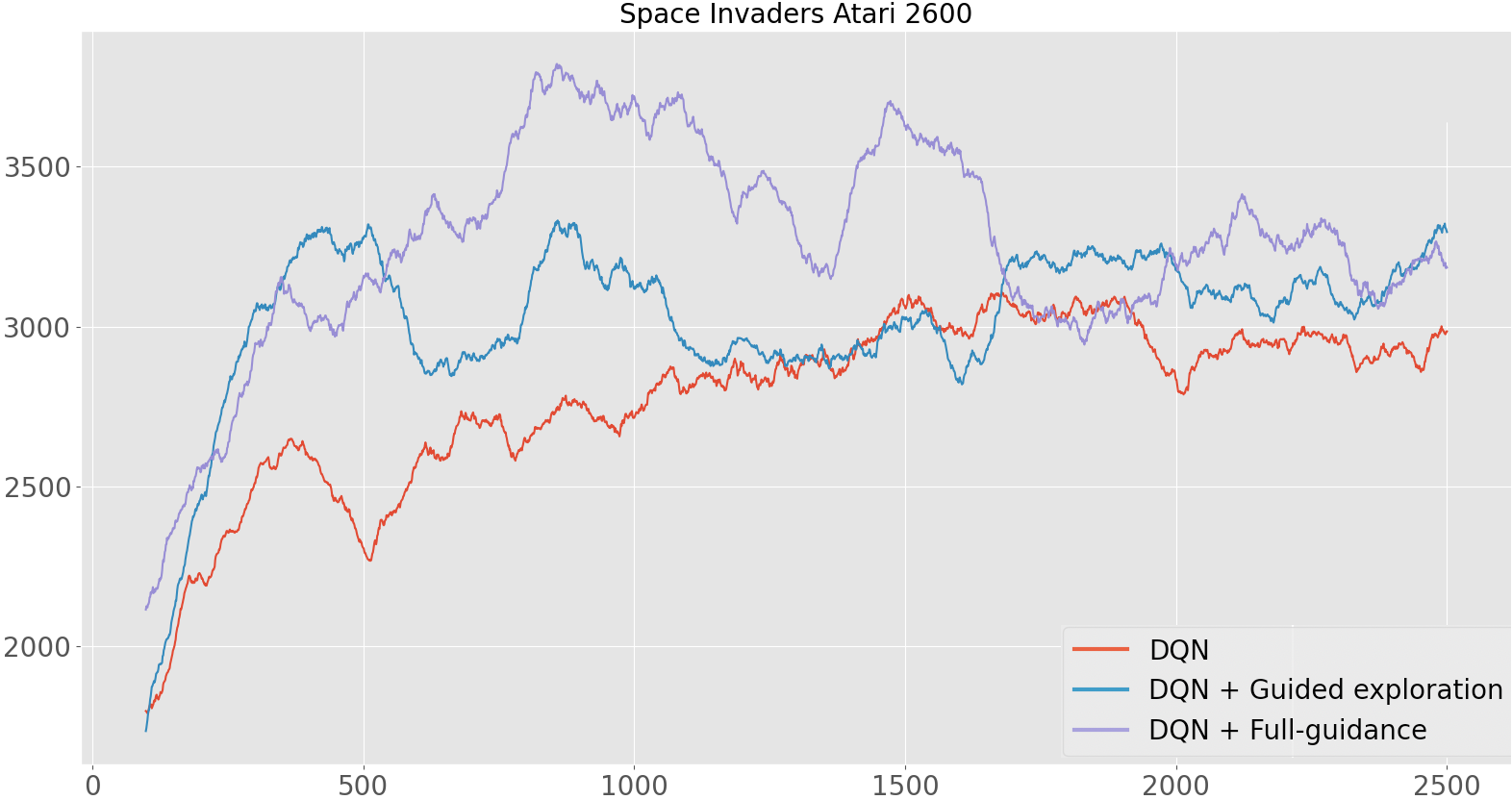}
\caption{Average rewards of the agents in five testing domains. }
\label{fig:learning-curves}
\end{figure*}

\paragraph{Improved Efficiency}
As we mentioned previously, we hypothesize that by preventing the agent from exploring unsafe actions, we could decrease the size of the search space thus increasing convergence speed. Figure \ref{fig:rel_performance} demonstrates how many episodes it took the safe $\epsilon$-greedy agents to achieve maximum performance of the base model. It demonstrates that for all domains the efficiency has improved significantly, achieving good performance faster than the base model. Figure \ref{fig:learning-curves} demonstrates average rewards achieved by agents in all domains. Safe $\epsilon$-greedy policy provided the agents with the initial jump-start in the performance and, in some cases, increased the convergence speed. Here we note that for Crossroad, vanilla DQN failed to learn after 25,000 episodes, whereas both DQN+GE and DQN+FG have converged to the optimal performance. The initial jump-start resulted in both agents surpassing vanilla DQN performance in the first episode.

\begin{table*}[t]
\centering
\begin{tabular}{|c|c|c|c|c|c|}
\hline
Agent & Crossroad & Freeway & Frostbite & Ms-Pacman & Space Invaders \\
\hline
DQN & -0.8 & 27.3 & 358.0 & 2780.0 & 1155.0 \\ 
DQN+GE & 1.0 & 28.9 & \textbf{490.0} & 3558.0 & 1470.0 \\ %
DQN+FG & \textbf{1.0} & \textbf{30.5} & 470.0 & \textbf{3820.0} & \textbf{1670.0} \\ 
Random & -1.0 & 0 & 65.2 & 307.3 & 148.0 \\
image-DQN & - & 30.3 & 328.3 & 2311.0 & 1449.7 \\
\hline
Human & 1.0 & 29.6 & 4334.7 & 6951.6 & 1668.7\\
\hline
\end{tabular}
\caption{The maximum reward achieved after training for 25,000 episodes (\textless 20M frames). Here image-DQN are the results of image-based DQN reported in \cite{mnih2015humanlevel}.}
\label{tab: atari_results}
\end{table*}

\paragraph{Better Performance}
In addition to bringing safety and efficiency, a safe $\epsilon$-greedy policy also improves the performance of the agents, resulting in achieving a higher maximum reward. Table \ref{tab: atari_results} demonstrates the maximum rewards achieved by the agents during training in all testing domains. We compare the achieved maximum performance of the agents to humans and a random agent. We also provide the results of the image-based DQN \cite{mnih2015humanlevel} to compare the performance impact of two different representations. In all domains safe $\epsilon$-greedy agents achieved better performance than the base models. We hypothesize that this is likely due to the two main reasons: increased efficiency and initial jump-start in the performance, and chosen testing domains. The first reason is rather self-explanatory, as better efficiency would result in achieving better performance faster, and an initial jump-start would provide a better starting point for the learning. Earlier, we mentioned that we selected the domains that require an agent to behave safely in order to achieve good performance. We hypothesize, that in comparison to some other domains, here safety correlates with the performance. Since the agent explores only safe states, it finds better policy faster.

\section{Conclusions and Limitations}
In this work, we proposed to use domain-specific symbolic safety rules to prevent agents from performing unsafe actions. We proposed a new safe $\epsilon$-greedy algorithm that injects symbolic rules into the training process of the reinforcement learning agent. We presented two versions of this algorithm that either completely restricts the agent in its actions during the entire training process, or only during the exploration of the world. We demonstrated how the safety rules can be defined for five different domains. In our experiments, we showed that preventing an agent from performing unsafe actions can significantly increase the efficiency of the agent, boost overall performance and make the training process much safer. We provided an additional experiment to show how eliminating unsafe actions completely results in much better learning efficiency rather than simply giving a negative reward. Finally, we addressed a known limitation and showed that safety rules can be learned and integrated into the training using our method.

One obvious limitation of using symbolic safety rules is that in more complex domains rules can be much more complicated and hard to define. However, as mentioned in the "Learning Rules Experiment" section, we hypothesize that in such complicated domains the rules can be learned by the agent using known rule-learning techniques. Since this goes beyond this paper, we leave it for future work.

Another limitation is that our method does not completely eliminate unsafe actions (Figure \ref{fig:all_bumps}). This is mainly due to the short horizon of defined symbolic rules. The rules that we use are only able to model one step ahead. This leaves a number of situations where inevitably agent has to perform some unsafe action. For example, in Freeway, there could be situations where the agent does not have enough time or any way to avoid colliding with a car. To avoid such scenarios more sophisticated and complex rules are needed. We hypothesize that to avoid unsafe actions completely a model may be needed, which essentially leads to a realm of planning. Another way could be to use a neurosymbolic approach. However, this discussion goes beyond this paper and we leave this to future work.

In conclusion, safe-$\epsilon$ greedy policy provides a flexible way to inject safety rules into the learning process to improve the safety and efficiency of the reinforcement learning agents.

\end{document}